\documentclass[11pt]{article}

\usepackage[preprint]{acl}
\usepackage{tikz}
\usetikzlibrary{positioning,arrows.meta,shadows,shapes.geometric,shapes.misc,calc,fit,backgrounds,
                shapes.symbols,decorations.pathreplacing}
\usepackage{tcolorbox}
\tcbuselibrary{breakable}
\usepackage{booktabs}
\usepackage{enumitem}
\usepackage{listings}
\usepackage{pifont}
\usepackage{tabularx}
\usepackage{svg}
\usepackage{amsfonts}
\usepackage{soul}
\usepackage{hyperref}
\usepackage{amssymb}
\newcommand{\cmark}{\ding{51}}
\newcommand{\xmark}{\ding{55}}

\lstdefinelanguage{json}{
    basicstyle=\ttfamily\footnotesize,
    numbers=none,
    breaklines=true,
    showstringspaces=false,
    stringstyle=\color{blue},
}
\usepackage{times}
\usepackage{latexsym}
\usepackage{subcaption}
\usepackage[T1]{fontenc}

\usepackage[utf8]{inputenc}
\usepackage{amsmath}

\usepackage{microtype}

\usepackage{inconsolata}

\usepackage{graphicx}
\usepackage{algorithmicx}
\usepackage{algcompatible}
\usepackage{algorithm}
\usepackage{algpseudocode}
\usepackage{booktabs}
\usepackage{longtable}
\usepackage[table]{xcolor}
\usepackage{caption}

\title{AMNESIA: A Large Scale Medical Unlearning Benchmark Suite with Disease-Informed Analysis}

\author{
 \textbf{Saeedeh Davoudi\thanks{Equal contribution.}\textsuperscript{1}},
 \textbf{Reihaneh Iranmanesh\footnotemark[1]\textsuperscript{1}},
 \textbf{Ophir Frieder\textsuperscript{1}},
 \textbf{Nazli Goharian\textsuperscript{1}}
\\
\\
 \textsuperscript{1}IR Lab, Computer Science Department, Georgetown University, Washington D.C.
\\
  \{saeedeh, rei, ophir, nazli\}@ir.cs.georgetown.edu}

\begin{document}
\maketitle
\begin{abstract}

Medical knowledge is continuously evolving. This creates a need to update or selectively forget information encoded in already-trained medical LLMs. Machine unlearning aims to remove the influence of specific training data from a model without full retraining. Yet, existing unlearning benchmarks rely on synthetic or small-scale general data, leaving clinical unlearning understudied. We introduce \textbf{AMNESIA}, the first large-scale, open source benchmark for medical unlearning, with 70,560 question-answer pairs from 8,820 patient notes across 11 disease categories. AMNESIA includes both \textit{factual} questions testing direct recall and \textit{reasoning} questions testing clinical inference. We use it to evaluate four widely used unlearning methods at both random patient and disease-level, and introduce a new metric for detecting leakage of medical terminology. We show that unlearning individual patients erodes knowledge of others with the same condition, calling for methods that can better separate patients from shared clinical knowledge.

\end{abstract}

\section{Introduction}

Machine learning models trained on sensitive medical data face a critical challenge: how to remove the influence of specific training examples while preserving overall model utility. This capability, known as \textit{machine unlearning}, is essential for healthcare applications where patients may request data deletion under privacy regulations like GDPR's ``right to be forgotten''~\citep{gdpr2016}, or where erroneous or outdated medical records must be excised from trained models without costly full retraining. Despite growing interest in machine unlearning across various domains, the medical field lacks standardized benchmarks for evaluating unlearning methods for clinical question-answering tasks.

To address these gaps, we introduce \textbf{AMNESIA}: \textbf{A} Large Scale \textbf{M}edical U\textbf{N}learning B\textbf{E}nchmark \textbf{S}uite with Disease-\textbf{I}nformed \textbf{A}nalysis. AMNESIA provides the first large-scale benchmark for evaluating machine unlearning in medical question-answering (QA), comprising 70,560 question-answer pairs derived from 8,820 patient notes across 11 disease categories. Unlike prior unlearning benchmarks built on synthetic or general-domain QAs (Appendix \ref{app:prior_work}), AMNESIA is grounded in real patient notes, spanning thousands of patients across many disease conditions. This is essential in medical contexts, where consent withdrawals, diagnostic revisions, and record corrections continually generate new unlearning targets. AMNESIA also distinguishes \textit{factual} questions (direct recall) from \textit{reasoning} questions (clinical inference) and supports both \textit{random patient-level} and \textit{disease-level} forgetting scenarios, enabling researchers to test whether models can forget patients with particular medical conditions.

Our contributions are as follows:

\begin{itemize}[leftmargin=*,noitemsep,topsep=2pt]
\item \textbf{First large-scale clinically-grounded unlearning benchmark:} 70,560 Question-Answer pairs (QAs) from 8,820 patient notes across 11 disease categories. Each patient has 4 factual (recall) and 4 reasoning (inference) QAs. \footnote{All data, code, and baselines will be publicly available without restrictions or data usage agreements.}
\item \textbf{Multi-level evaluation across diverse forget/retain splits:} Evaluation across random patient-level and disease-level forget/retain splits.
\item \textbf{Comprehensive unlearning analysis:} Four representative unlearning methods are evaluated and analyzed.
\item \textbf{Medically-aware unlearning evaluation:} Novel disease-focused keyword evaluation for forget performance.
\item \textbf{Release of medical models and dataset:} Our models are publicly available. Our detailed data construction pipeline makes AMNESIA reproducible and easy to extend.
\end{itemize}

\begin{figure*}[h]
    \centering
\includegraphics[width=\textwidth]{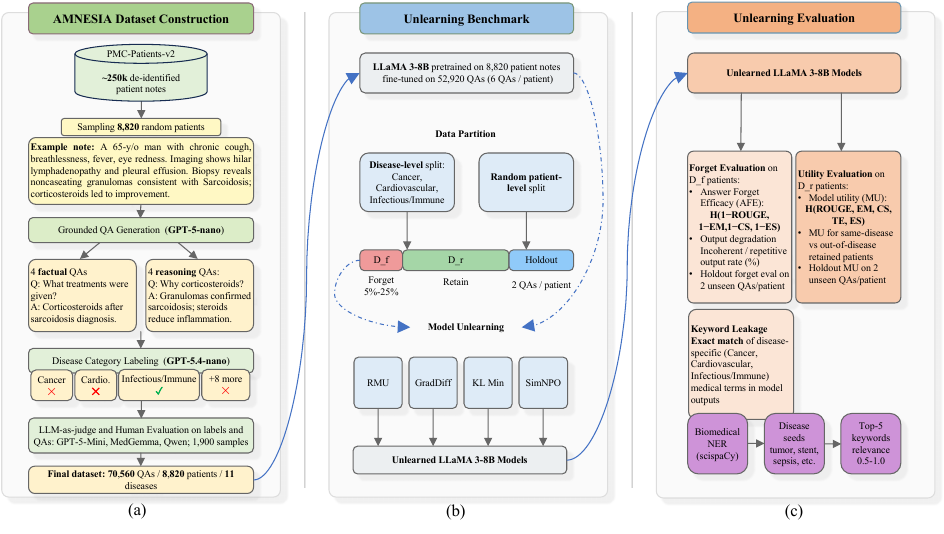}
    \caption{AMNESIA Dataset Construction (a), Unlearning Benchmark (b), and Evaluation Pipeline (c)}
    \label{fig:amnesia-overview}
\end{figure*}

\section{Related Work}

\paragraph{Machine Unlearning Benchmarks.}
Machine unlearning aims to remove the influence of specific training data from a model without full retraining~\citep{cao2015towards}.
TOFU~\citep{maini2024tofu} establishes the standard benchmark for LLM unlearning using synthetic author profiles, with methods including Gradient Difference~\citep{liu2022continual}, KL Minimization~\citep{nguyen2020variational}, and Negative Preference Optimization~\citep{zhang2024negative}. R-TOFU~\citep{yoon2025rtofu} extends TOFU to Large Reasoning Models, showing that answer-only objectives leave residual forget traces in chain-of-thought reasoning.
MUSE~\citep{shi2025muse} evaluates forgetting in LLMs but does not address clinical or cross-modal structures.
Recent work further shows that existing methods fail on structured, multi-hop knowledge~\citep{choi2024breaking}, motivating more realistic evaluation settings. We follow OpenUnlearning~\citep{dorna2026openunlearning}, a unified framework integrating 13 unlearning algorithms and 16 evaluations across TOFU, MUSE, and WMDP \citep{li2024wmdp}, to inform our selection and standardized comparison of unlearning methods.

\paragraph{Medical Question Answering.}
Clinical NLP has benefited from large-scale datasets such as MIMIC-CXR ~\citep{johnson2019mimic}, with models like LLaVA-Med~\citep{li2023llavamed} and Asclepius~\citep{kweon2023publicly} demonstrating that LLMs fine-tuned on patient notes acquire patient-specific knowledge.
Recent benchmarks have pushed medical QA in several directions: evaluating LLM reasoning and explanation on challenging clinical cases~\citep{chen2025benchmarking}, detecting hallucinations in medical outputs~\citep{pandit2025medhallu}, assessing multi-hop biomedical reasoning across knowledge graphs~\citep{kim2025biohop}, and supporting argumentative explanation of diagnoses~\citep{sviridova2024casimedicos}. However, these benchmarks do not address machine unlearning in a medical QA setting.

\paragraph{Machine Unlearning in Medical AI.}
Privacy regulations, including HIPAA~\citep{hipaa1996} and GDPR~\citep{gdpr2016}, motivate the selective removal of patient data from trained models, with prior defenses such as federated learning~\citep{mcmahan2017communication} offering only partial protection. Other work has explored unlearning multimodal patient information in clinical imaging settings~\citep{hardan2025forgetmi}.
MedForget~\citep{wu2025medforget} is the closest prior work, providing a hierarchical multimodal benchmark of 3,840 Visual Question Answering (VQA) pairs across 64 patients from MIMIC-CXR.
In contrast, AMNESIA scales to 70,560 clinical note–grounded QAs and introduces disease-level unlearning, a clinically motivated setting not addressed by prior work.

\section{AMNESIA Dataset}
 Figure~\ref{fig:amnesia-overview} illustrates the AMNESIA framework. Given a patient note, AMNESIA includes factual and reasoning questions, each with gold-standard answers. After unlearning, models should produce substantially different (uninformative) responses to questions about forgotten patients while maintaining accurate answers for retained patients.
 
\subsection{Data Construction (Figure \ref{fig:amnesia-overview}.a)}

\paragraph{Dataset Selection and Sampling}
We use PMC-Patients-v2~\citep{zhao2023pmc}, a large-scale publicly available dataset of de-identified patient notes extracted from reports in PubMed Central~\citep{roberts2001pmc}. From this corpus, we randomly sampled 8,820 patient notes to create a computationally feasible yet substantial dataset for our study.

\paragraph{Disease Category Assignment}
We used \textsc{GPT-5.4-nano} to assign a disease from Centers for Disease Control and Prevention (CDC)'s classification of common diseases \footnote{\url{https://www.cdc.gov/nchs/fastats/diseases-and-conditions.htm}} (11 diseases listed in Table \ref{tab:dataset-stats}) to each patient based on its note title. The note title is a single sentence describing the patient note. The prompt is designed to enforce \textsc{GPT-5.4-nano} to map each note's title to a single disease category (see Appendix~\ref{app:disease-prompt}). This categorization enables AMNESIA's disease-informed analysis of machine unlearning methods. We evaluate these categories in the \nameref{par:human-eval}.

\paragraph{Question-Answer Generation}
For each patient, we use \textsc{GPT-5-nano} to generate 8 QAs: 4 factual QAs testing direct information recall from the patient's note and 4 reasoning QAs requiring inference across multiple findings within the note, making them more challenging than factual questions. The generation prompt (Appendix~\ref{app:qa-prompt}) enforces strict quality criteria: questions must be answerable only from the specific patient's details (not general medical knowledge), must avoid any personal information, and must maintain clinical specificity. This generation strategy produced 70,560 total QAs (35,280 factual and 35,280 reasoning), ensuring coverage of both knowledge retrieval and clinical reasoning capabilities. Table~\ref{tab:dataset-stats} presents the final dataset composition across disease categories.
\begin{table}[h]
\centering
\fontsize{9}{9}\selectfont
\setlength{\tabcolsep}{2.5pt}
\begin{tabularx}{\columnwidth}{@{}>{\raggedright\arraybackslash}Xrrrrr@{}}
\toprule
\textbf{Disease Category} & \textbf{\# Patients} & \textbf{\# QA} & \textbf{Min} & \textbf{Avg} & \textbf{Max} \\
\midrule
Cancer & 2,669 (30.3\%) & 21,352 & 55 & 695 & 3,920 \\
Infectious/Immune & 1,964 (22.3\%) & 15,712 & 59 & 769 & 4,216 \\
Cardiovascular & 1,426 (16.2\%) & 11,408 & 58 & 693 & 2,796 \\
Digestive and Liver & 590 (6.7\%) & 4,720 & 48 & 713 & 3,160 \\
Dementia and Mental Health & 490 (5.6\%) & 3,920 & 66 & 693 & 3,724 \\
Arthritis and Bone & 441 (5.0\%) & 3,528 & 67 & 700 & 2,323 \\
Respiratory and Allergies & 435 (4.9\%) & 3,480 & 42 & 707 & 2,430 \\
Kidney Disease & 350 (4.0\%) & 2,800 & 95 & 738 & 2,212 \\
Diabetes & 208 (2.4\%) & 1,664 & 92 & 736 & 3,489 \\
Oral and Dental Health & 192 (2.2\%) & 1,536 & 77 & 698 & 2,271 \\
Anemia or Iron Deficiency & 55 (0.6\%) & 440 & 178 & 740 & 1,488 \\
\midrule
\textbf{Total} & \textbf{8,820} & \textbf{70,560} & \textbf{42} & \textbf{716} & \textbf{4,216} \\
\bottomrule
\end{tabularx}
\caption{AMNESIA dataset statistics with disease categories. Min, Avg, and Max denote the shortest, average, and longest note sizes (in tokens) within each disease category.}
\label{tab:dataset-stats}
\end{table}

\paragraph{Human-in-the-Loop Evaluation}\label{par:human-eval}

We validate both disease labels and QAs under majority vote with a three-judge LLM panel: a medical-specialized model (\textsc{MedGemma-27B} \citep{medgemma2024}),
a general-purpose reasoning model (\textsc{Qwen3-32B} \citep{qwen3}),
and a frontier model (\textsc{GPT-5-mini} \citep{openai2025gpt5}). For disease labels, we sample 900 patient notes ($\sim$10\% of total notes) from
the three most populated diseases (300 each from Cancer,  Infectious/Immune, and Cardiovascular). For QAs, we sample 500 patients and take 1 factual and 1 reasoning QA from each, yielding
1{,}000 QAs. The judge panel supports $\mathbf{91.8\%}$ of disease labels and rates $\mathbf{97.6\%}$ of QA triples as valid against four criteria (clarity, correctness, derivability, clinical meaning); a biology student's annotations of the same patient notes agree with the judges' majority vote on $\mathbf{90.4\%}$ of disease labels. Together, these results indicate that AMNESIA's QAs and disease labels are trustworthy for unlearning evaluation. Full protocols, prompts, and per-judge breakdowns are in Appendices~\ref{app:human-validation} and~\ref{app:qa-validity}.

\subsection{Data Partition (Figure \ref{fig:amnesia-overview}.b)}
\label{sec:data-partition}
Our partition strategy creates hierarchical forget/retain splits at two levels: random patient-level and disease-level.

\paragraph{Random Patient-Level Splits}
Similar to existing unlearning benchmarks~\citep{maini2024tofu,wu2025medforget}, we create forget splits by randomly selecting patients at 5\%, 10\%, 15\%, 20\%, and 25\% of the total number of patients, where each smaller forget set is a proper subset of the next larger one (i.e., forget\_5 $\subset$ forget\_10 $\subset$ forget\_15 $\subset$ forget\_20 $\subset$ forget\_25). For each forget split, the corresponding retain split contains all remaining patients.

\paragraph{Disease-Level Splits}
We build disease-level splits for the three most prominent diseases in our dataset: Cancer, Infectious/Immune, and Cardiovascular. Each disease-level split matches the size of the corresponding random patient-level split, sampled only from the target disease. The size of the largest possible forget set for each disease depends on the total number of patients having that disease in the dataset (Table~\ref{tab:dataset-stats}):
\begin{itemize}[noitemsep,topsep=2pt,leftmargin=*]
\item \textbf{Cancer (30.3\% of all patients): } 5\%, 10\%, 15\%, 20\%, 25\% forget splits.
\item \textbf{Infectious/Immune Diseases (22.3\% of all patients):} 5\%, 10\%, 15\%, 20\% forget splits.
\item \textbf{Cardiovascular (16.2\% of all patients):} 5\%, 10\%, 15\% forget splits.
\end{itemize}

Each disease-level split maintains two key properties: (1) the subset relationship holds within each disease (e.g., cancer\_forget\_5 $\subset$ ... $\subset$ cancer\_forget\_25), and (2) all patients with a given disease label that appear in the random patient forget split are also included in the corresponding disease-level forget split at the same percentage (e.g., all cancer patients in forget\_5 are guaranteed to appear in cancer\_forget\_5. Appendix \ref{app:data-statistics} shows the size of each data split.).  The 900 samples selected for human-in-the-loop evaluation of disease categories are drawn from cancer\_forget\_5, cardio\_forget\_5, and infectious/immune\_forget\_5 splits, which are shared among all forget sets of that disease (See \nameref{par:human-eval}).

\paragraph{Holdout Dataset}
To evaluate generalization, we create a holdout set by withholding a factual and a reasoning QA per patient (17,640 total), never seen during fine-tuning and unlearning.  Complete statistics for all splits are provided in Appendix~\ref{app:data-statistics}.

\section{Unlearning Benchmark (Figure \ref{fig:amnesia-overview}.b)}
\label{sec:benchmark}

We establish a medical unlearning benchmark through a four-stage workflow: (1) Pre-training on patient notes, (2) instruction fine-tuning on QAs, (3) applying unlearning methods on the fine-tuned model, and (4) evaluation across multiple metrics.

\subsection{Base Language Model}
\label{sec:base-model}

We select \textsc{LLaMA 3-8B} as the base language model, following established practices in medical AI~\citep{kweon2023publicly}. We perform continued pre-training on 8,820 patient notes using standard next-token prediction, followed by instruction fine-tuning on 52,920 QAs (see Appendix~\ref{app:training-hyperparams} for hyperparameter details). The resulting fine-tuned model serves as the baseline for all unlearning experiments.

\subsection{Unlearning Methods}
\label{sec:unlearning-methods}

We evaluate four top-performing unlearning methods from OpenUnlearning~\citep{dorna2026openunlearning}, covering diverse algorithmic families. All methods operate on the forget set $\mathcal{D}_f$ (patients to unlearn) and the retain set $\mathcal{D}_r$ (patients to preserve), applied to the fine-tuned model. Following the instruction fine-tuning configuration, all unlearning methods are trained with forget/retain patient notes and their QAs as input:

 \textbf{RMU}~\citep{li2024wmdp} steers internal representations of forget set inputs toward random targets at intermediate layers while preserving retain set representations, avoiding output-level gradient manipulation. In our experiments, steering is applied at layer~7, consistent with the OpenUnlearning default values.

\textbf{GradDiff}~\citep{liu2022continual} jointly maximizes the loss on $\mathcal{D}_f$ and minimizes the loss on $\mathcal{D}_r$ via the objective $\mathcal{L}_{\text{GD}} = -\mathcal{L}(\mathcal{D}_f, w) + \mathcal{L}(\mathcal{D}_r, w)$, where $\mathcal{L}(\mathcal{D}_f, w)$ is implemented as gradient ascent on the forget set negative log-likelihood. Rather than fine-tuning the full model, we perform \emph{layer-selective} unlearning: all parameters are frozen except those in layer 7. \textbf{Freezing all other layers limits parameter drift on $\mathcal{D}_r$, reduces catastrophic forgetting and training cost.}

\textbf{KL-Min(imization)}~\citep{nguyen2020variational} uses the same GradDiff forget objective and the same layer-7 parameter mask. KL-Min applies gradient ascent on $\mathcal{D}_f$ while regularizing $\mathcal{D}_r$ outputs to match the original model via KL divergence, preventing model collapse.

\textbf{SimNPO}~\citep{fan2024simnpo} treats forgotten samples as negative preferences without reference-model dependency, addressing bias issues in standard NPO~\citep{zhang2024negative}. Like GradDiff and KL-Min, we update only layer~7.

\section{Evaluation Metrics (Figure \ref{fig:amnesia-overview}.c)}
To assess the trade-off between knowledge removal and utility preservation, we use four metrics from \citep{yoon2025rtofu} and one from \citep{dorna2026openunlearning}. We also introduce a novel disease-informed leakage evaluation metric.

\subsection{Disease Keyword Leakage Evaluation}
\label{sec:keyword-leakage}
 
Standard unlearning metrics measure semantic and lexical similarity to references, but do not test whether disease-specific terminology about forget patients still surfaces in outputs. We therefore introduce a leakage metric based on disease-specific token matching.
 
\paragraph{Keyword Extraction} 
For each QA instance, we build a disease-specific keyword set from the note, question, and answer. We parse each field with scispaCy~\citep{neumann2019scispacy} and collect biomedical named entities and noun chunks, dropping generic clinical terms (e.g., \emph{patient}, \emph{history}). Relevance to diseases is scored lexically against curated seed lists for Cancer (e.g., \emph{chemotherapy}, \emph{metastasis}), Infectious/Immune (e.g., \emph{sepsis}, \emph{autoimmune}), and Cardiovascular (e.g., \emph{stent}, \emph{troponin}). Complete seed lists and generic terms are detailed in Appendix~\ref{app:ls}.
 
For a candidate phrase $p$ and seed $s$, we assign lexical relevance $\mathrm{rel}(p)\in[0,1]$ using exact or partial phrase matching. Each candidate receives a weight $s_i$ based on its frequency in that field. We then merge the top K keywords from the notes and all surviving keywords from QAs.
 
\paragraph{Leakage Score (LS)} 
Let $y$ be the model's generated answer. Leakage is the weighted fraction of keywords that reappear in $y$ as a contiguous token subsequence,
\begin{equation}
\mathrm{LS}(y)=
\frac{\sum_i s_i \cdot \mathbb{1}\!\left[k_i \in_{\mathrm{tok}} y\right]}
{\sum_i s_i}
\in [0,1],
\end{equation}
where $\mathbb{1}[k_i \in_{\mathrm{tok}} y]$ is $1$ if the keyword $k_i$ occurs consecutively in $y$ after tokenization (case-insensitive, alphanumeric only). We report $\mathrm{LS}$ as a percentage (mean over forget set QAs with at least one keyword). Higher values indicate more reproduction of disease-specific terms from the note.

\subsection{Established Metrics}
\paragraph{Cosine Similarity (CS)} We measure semantic similarity using PubMedBERT embeddings \citep{pubmedbert2020}.

\paragraph{Entailment Score (ES)} We employ PubMedBERT-MNLI-MedNLI \citep{deka-nli} to verify factual consistency between model responses and reference answers.

\paragraph{Syntactic Overlap} We calculate \textbf{ROUGE-L recall} and position-aligned token \textbf{Exact Memorization (EM)} \citep{dorna2026openunlearning} to track lexical retention.

\paragraph{Token Entropy (TE)} We monitor normalized unigram entropy \citep{yoon2025rtofu} to identify instances of model collapse or repetitive output (higher indicates more diverse output).

We utilize the \textbf{harmonic mean} ($H$) to compute primary performance indices. This ensures that a model cannot achieve high scores by over-optimizing for a single metric while failing in others:

\paragraph{Model Utility (MU)} Measures the preservation of medical knowledge in the retain set,
\begin{equation}
MU = H(\text{ROUGE, EM, CS, TE, ES})
\end{equation}

\paragraph{Answer Forget Efficacy (AFE)} Quantifies unlearning success on the forget set by aggregating the inversion of quality metrics ($1 - \text{score}$),
\begin{equation}
AFE = H(1-\text{ROUGE, } 1-\text{EM, } 1-\text{CS, } 1-\text{ES})
\end{equation}

Following \citep{yoon2025rtofu}, we exclude TE from this aggregate since it measures how repetitive the text is, which is unrelated to whether facts were actually forgotten. Instead, we report \textbf{degradation score}.

\paragraph{Degradation Score (DS)}
To detect gibberish output, we report a DS,

\begin{equation}
\text{DS} = \frac{|\{x \in \mathcal{D}_f : TE(x) = 0\}|}{|\mathcal{D}_f|},
\end{equation}
which is the fraction of forget set generations with zero mean TE. These correspond to fully repetitive token outputs; a high DS indicates output collapse rather than knowledge removal (Table \ref{tab:keyword_leakage}).

\section{Results and Analysis}
We evaluate 4 unlearning methods across two patient selection strategies: random patient-level and disease-level. For all experiments, MU is measured on the retain set, and AFE is measured on the forget set. We expect the MU for unlearned models to be close to the baseline MU, and the AFE to exceed the baseline AFE. MU of the holdout set is measured on the unseen QAs of retain patients, and AFE is measured on unseen QAs of forget patients. We report results separately for factual and reasoning QAs.

\subsection{Do Models Forget Random Patients Regardless of Their Disease Profile?}
\label{subsec:6.1}

\begin{figure}[h]
    \centering
    \includegraphics[width=\columnwidth, height= 11.3cm]{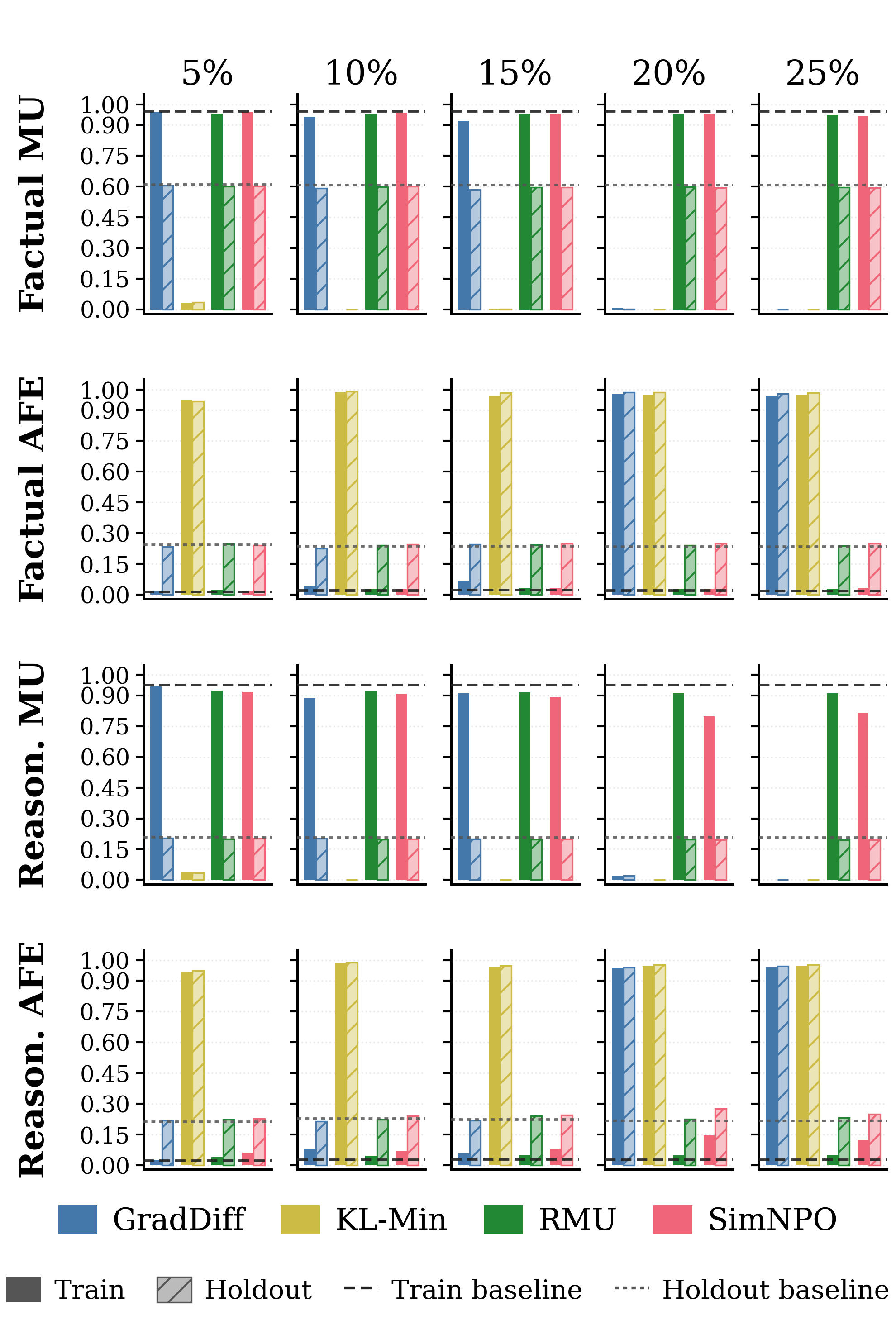}
    \caption{Unlearning performance at random patient-level. Rows correspond to MU and AFE, each measured on factual (top two rows) and reasoning (bottom two rows) QAs. Columns
are forget set split sizes. For every method, the solid bar reports performance on train data, hatched bar on holdout data. Dashed lines are baseline (finetuned model) results on train and holdout datasets.}
    \label{fig:model_performance_patient}
\end{figure}

Figure~\ref{fig:model_performance_patient} presents random patient-level results, examining whether unlearning methods can forget patient information across different forget set sizes from 5\% to 25\%. On training (seen) questions (solid bars), RMU and SimNPO maintain high MU (above 0.9 in factual and above 0.75 in reasoning QAs) and low AFE (< 0.15) across all forget set sizes, \textbf{indicating ineffective unlearning in the random patient-level setting.} KL-Min exhibits the opposite pattern: high AFE (above 0.9 for both QA types) but low MU (${\sim}0.0$), achieving forgetting by generating repeated tokens (Table ~\ref{tab:keyword_leakage}) rather than selective erasure. GradDiff balances both metrics at 5--15\% splits (MU ${\sim}0.9$ for both QA types) but collapses at 20--25\%, similar to KL-Min, losing model utility as more patients are forgotten. 

Holdout QAs (hatched bars) yield much lower MU baseline values than train QAs (${\sim}0.6$ vs  ${\sim}1.00$ for factual and ${\sim}0.2$ vs ${\sim}0.95$ for reasoning) since these QAs are unseen. \textbf{The large gap between factual and reasoning holdout MU values suggests that unlearned models recall direct facts from the note but fail to reason from it.} Holdout AFE also remains low for RMU and SimNPO in both factual and reasoning QAs and close to baseline. This suggests that models struggle with novel QAs, creating a false sense of forgetting. Also, it means that \textbf{unlearned models forget specific QAs from training data rather than truly erasing learned patient information (notes), which remains accessible via alternative questions.}

\begin{figure*}[h!]
    \centering
    \includegraphics[width=\textwidth, height=11cm]{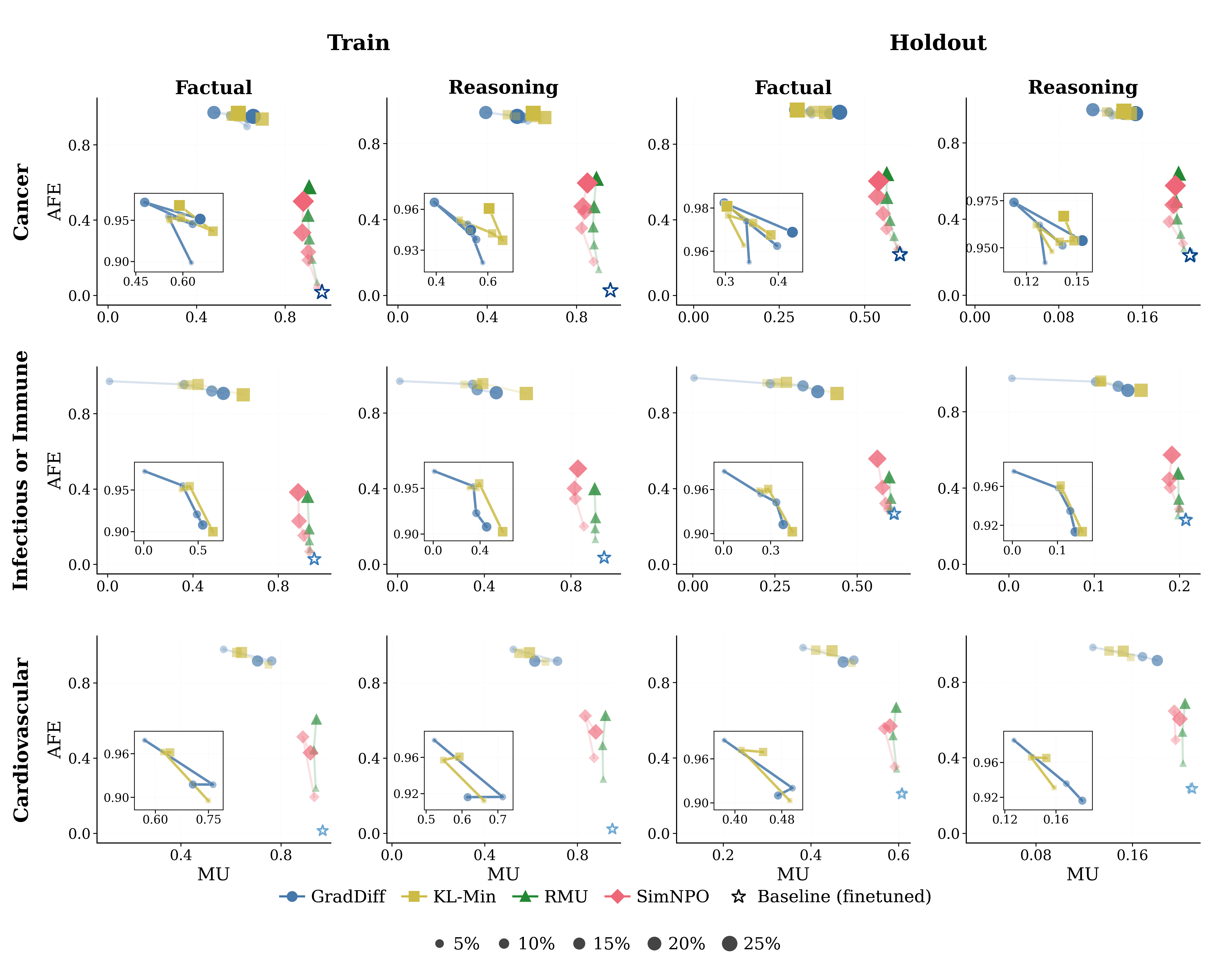}
    \caption{Unlearning performance at disease-level. Horizontal axis is MU on  \emph{retain} patients; vertical axis is AFE on \emph{forget} patients. Each colored trace is one unlearning method and stars are baseline values (finetuned model before unlearning); marker opacity scales with the forget set size ($5\%$--$25\%$ splits). Light polylines connect successive forget splits. Each inset zooms the region where GradDiff and KL-Min traces concentrate.}
    \label{fig:model_performance_disease}
\end{figure*}

\begin{figure}[!t]
    \centering
    \begin{subfigure}[b]{\linewidth}
        \includegraphics[width=\linewidth, height=4.5cm]{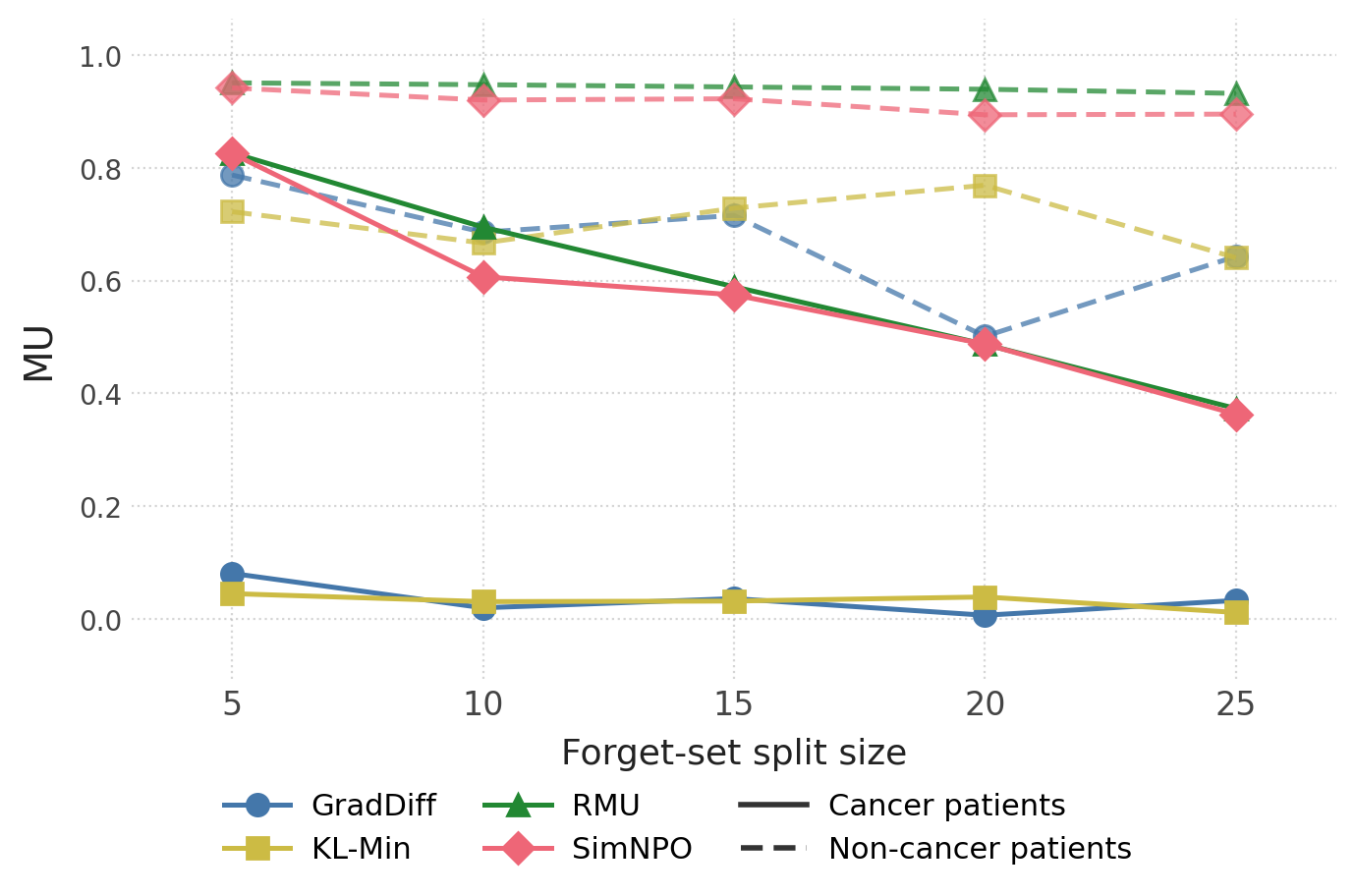}
        \caption{Cancer vs.\ non-cancer retain patients.}
        \label{fig:mu_cancer_overlay}
    \end{subfigure}
    \begin{subfigure}[b]{\linewidth}
        \includegraphics[width=\linewidth, height=4.5cm]{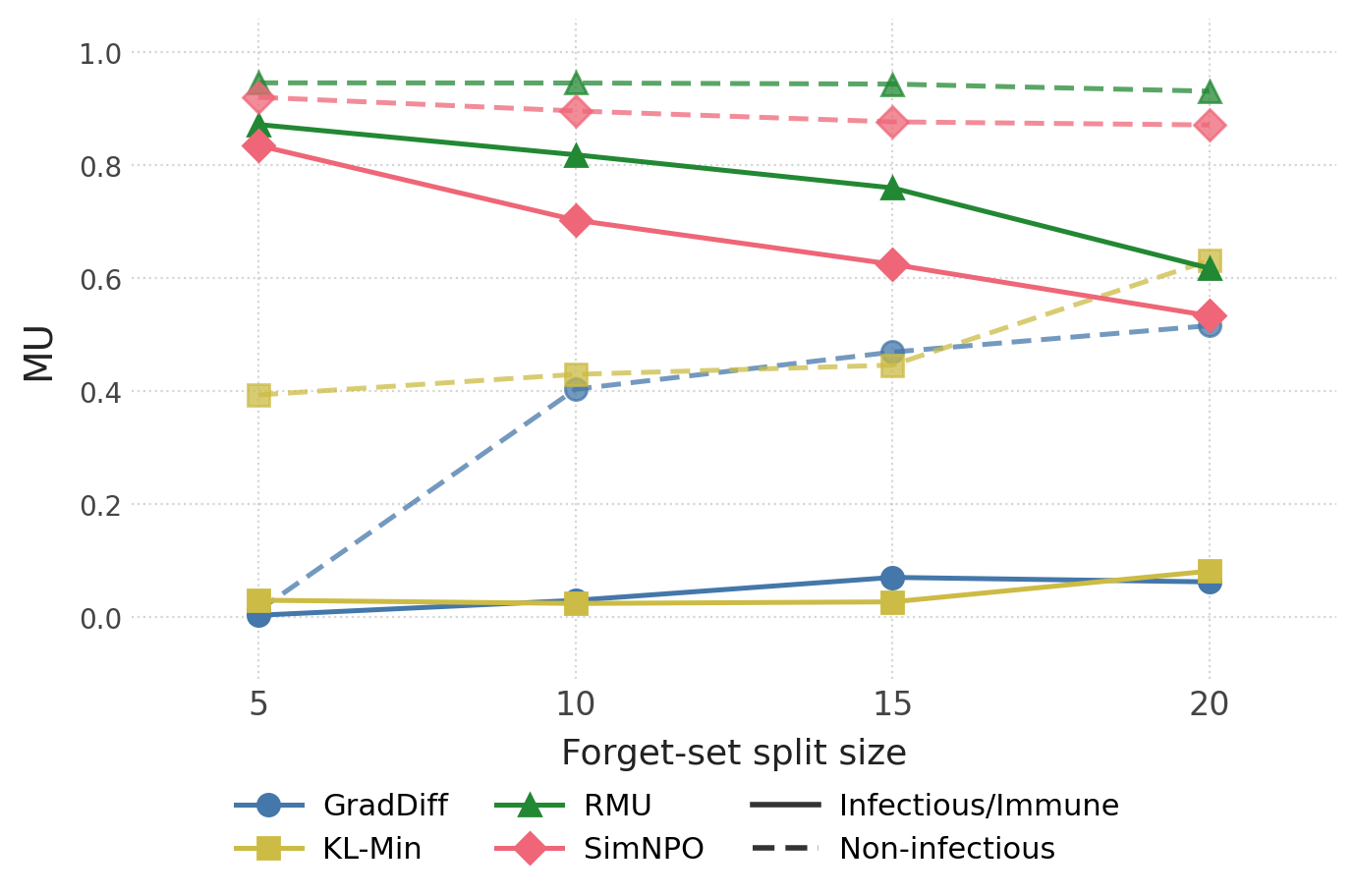}
        \caption{Infectious/Immune vs.\ non-infectious retain patients.}
        \label{fig:mu_immune_overlay}
    \end{subfigure}
    \caption{Retain MU on seen questions for same-disease (solid) vs.\ out-of-disease (dashed)
    patients across forget set sizes for cancer and infectious/immune cohorts, averaged across \textbf{both} factual and reasoning questions.}
    \label{fig:mu_overlay}
\end{figure}

\begin{table*}[t]
\centering
\small
\setlength{\tabcolsep}{3pt}
\resizebox{\linewidth}{!}{%
\begin{tabular}{ll ccccc}
\toprule
 & & \multicolumn{5}{c}{\textbf{Forget Set LS and DS (\%) (Factual / Reasoning)}} \\
\cmidrule(lr){3-7}
\textbf{Disease} & \textbf{Method} & \textbf{5\%} & \textbf{10\%} & \textbf{15\%} & \textbf{20\%} & \textbf{25\%} \\
\midrule
\textbf{Cancer} & GradDiff
  & 0.5 (76.2) / \textbf{0.5} (75.0)
  & 0.3 \underline{(93.8)} / \textbf{0.3} \underline{(93.2)}
  & 0.3 (90.9) / \textbf{0.3} \underline{(89.5)}
  & \textbf{0.1} \underline{(97.7)} / \textbf{0.1} \underline{(97.3)}
  & 0.2 (87.5) / 0.3 (86.8) \\
 & KL-Min
  & \textbf{0.2} \underline{(86.4)} / \textbf{0.5} \underline{(81.3)}
  & \textbf{0.2} (91.0) / \textbf{0.3} (92.1)
  & \textbf{0.2} \underline{(91.6)} / \textbf{0.3} (87.3)
  & 0.2 (90.9) / 0.3 (89.9)
  & \textbf{0.1} \underline{(95.4)} / \textbf{0.1} \underline{(94.3)} \\
 & RMU
  & 13.9 \textbf{(0.3)} / \underline{24.6} \textbf{(0.0)}
  & \underline{13.9} \textbf{(0.2)} / \underline{22.8} \textbf{(0.0)}
  & \underline{14.0} (0.3) / \underline{20.9} \textbf{(0.0)}
  & \underline{14.2} \textbf{(0.1)} / \underline{18.6} \textbf{(0.0)}
  & 10.4 \textbf{(0.1)} / \underline{12.7} \textbf{(0.0)} \\
 & SimNPO
  & \underline{14.1} (0.5) / 24.1 \textbf{(0.0)}
  & 13.6 \textbf{(0.2)} / 16.9 \textbf{(0.0)}
  & 11.9 \textbf{(0.1)} / 14.5 \textbf{(0.0)}
  & 10.9 (0.3) / 13.8 \textbf{(0.0)}
  & \underline{12.1} \textbf{(0.1)} / 11.6 \textbf{(0.0)} \\
\midrule
\textbf{Infectious/} & GradDiff
  & \textbf{0.0} \underline{(100.0)} / \textbf{0.0} \underline{(100.0)}
  & 0.4 (92.1) / \textbf{0.6} (89.8)
  & 0.6 (82.1) / 0.9 (82.8)
  & \textbf{0.7} (72.2) / \textbf{1.0} (72.8)
  & -- / -- \\
 \textbf{Immune} & KL-Min
  & 0.3 (92.1) / 0.6 (87.2)
  & \textbf{0.3} \underline{(92.4)} / \textbf{0.6} \underline{(90.8)}
  & \textbf{0.3} \underline{(95.2)} / \textbf{0.4} \underline{(94.0)}
  & 1.2 \underline{(82.4)} / 1.3 \underline{(81.4)}
  & -- / -- \\
 & RMU
  & \underline{14.8} \textbf{(0.4)} / \underline{23.8} \textbf{(0.0)}
  & \underline{14.2} \textbf{(0.3)} / \underline{21.9} \textbf{(0.0)}
  & \underline{14.0} \textbf{(0.3)} / \underline{20.0} \textbf{(0.1)}
  & \underline{12.2} \textbf{(0.2)} / \underline{16.0} \textbf{(0.1)}
  & -- / -- \\
 & SimNPO
  & \underline{14.8} \textbf{(0.4)} / 21.3 \textbf{(0.0)}
  & \underline{14.2} \textbf{(0.3)} / 14.9 \textbf{(0.0)}
  & 12.3 (0.5) / 12.6 (0.2)
  & 10.9 \textbf{(0.2)} / 10.4 \textbf{(0.1)}
  & -- / -- \\
\midrule
\textbf{Cardio-} & GradDiff
  & \textbf{0.2} \underline{(97.7)} / \textbf{0.1} \underline{(97.6)}
  & 0.7 (79.6) / 0.7 (80.1)
  & 0.4 (88.9) / 0.5 (88.8)
  & -- / --
  & -- / -- \\
 \textbf{vascular} & KL-Min
  & 0.6 (7.0) / 0.5 (5.5)
  & \textbf{0.1} \underline{(97.3)} / \textbf{0.1} \underline{(97.5)}
  & \textbf{0.2} \underline{(97.2)} / \textbf{0.1} \underline{(97.3)}
  & -- / --
  & -- / -- \\
 & RMU
  & 11.3 (0.3) / \underline{14.8} \textbf{(0.2)}
  & 10.3 \textbf{(0.2)} / \underline{12.5} (0.1)
  & 5.9 \textbf{(0.1)} / 7.5 (0.1)
  & -- / --
  & -- / -- \\
 & SimNPO
  & \underline{11.9} \textbf{(0.2)} / 11.1 \textbf{(0.2)}
  & \underline{10.4} \textbf{(0.2)} / 8.7 \textbf{(0.0)}
  & \underline{8.8} \textbf{(0.1)} / \underline{8.5} \textbf{(0.0)}
  & -- / --
  & -- / -- \\
\midrule
\textbf{Random} & GradDiff
  & \underline{14.3} \textbf{(0.2)} / \underline{26.0} \textbf{(0.0)}
  & \underline{14.7} \textbf{(0.2)} / 25.4 \textbf{(0.0)}
  & 14.4 \textbf{(0.2)} / 25.5 \textbf{(0.0)}
  & 0.1 (21.1) / 0.3 (16.7)
  & \textbf{0.0} \underline{(100.0)} / \textbf{0.0} \underline{(100.0)} \\
\textbf{Patient} & KL-Min
  & \textbf{0.7} \underline{(64.2)} / \textbf{1.5} \underline{(33.4)}
  & \textbf{0.0} \underline{(98.1)} / \textbf{0.0} \underline{(98.4)}
  & \textbf{0.0} \underline{(96.7)} / \textbf{0.0} \underline{(97.4)}
  & \textbf{0.0} \underline{(99.4)} / \textbf{0.0} \underline{(99.4)}
  & \textbf{0.0} (99.7) / \textbf{0.0} (99.7) \\
 & RMU
  & \underline{14.3} \textbf{(0.2)} / 25.6 \textbf{(0.0)}
  & 14.6 (0.3) / \underline{26.0} \textbf{(0.0)}
  & \underline{14.6} (0.3) / \underline{25.7} \textbf{(0.0)}
  & \underline{14.6} \textbf{(0.4)} / \underline{25.5} \textbf{(0.0)}
  & \underline{14.5} \textbf{(0.4)} / \underline{25.3} \textbf{(0.0)} \\
 & SimNPO
  & \underline{14.3} \textbf{(0.2)} / 24.8 \textbf{(0.0)}
  & 14.5 (0.3) / 25.6 \textbf{(0.0)}
  & 14.5 (0.3) / 24.0 \textbf{(0.0)}
  & 14.5 \textbf{(0.4)} / 23.5 (0.1)
  & 14.4 \textbf{(0.4)} / 23.1 \textbf{(0.0)} \\
\bottomrule
\end{tabular}}
\caption{Forget set leakage score (LS) and degradation score (DS) in percentages across random patient and disease-level splits, unlearning methods, and forget set sizes. Each cell reports factual / reasoning LS with DS in parentheses; bold = lowest, underlined = highest across methods for each QA type.}
\label{tab:keyword_leakage}
\end{table*}

\subsection{Does Forgetting Patients with the Same Disease Differ from Forgetting Random Patients?}

Analysis in \ref{subsec:6.1} showed that model unlearning for forgetting random patients either collapses (high AFE and low MU) or is ineffective (high MU and low AFE). We now examine whether this behavior changes when forgotten patients share the same disease.

The evaluation of the disease-level (Figure~\ref{fig:model_performance_disease}) reveals a fundamentally different pattern from the random-patient unlearning. We plot forget set AFE versus retain set MU for the three most populated diseases (Cancer, Infectious/Immune, Cardiovascular) across different forget set sizes. Perfect unlearning appears in the top-right corner (high MU, high AFE). Unlike random patients, patients sharing the same disease are interconnected through shared medical characteristics. KL-Min and GradDiff behave similarly across all three diseases. Both achieve moderate MU ($\sim$0.4--0.7) while preserving high AFE ($\sim$0.90--0.97). The pattern is noisy and non-monotonic for both QA types.

For RMU and SimNPO, MU also remains high across all three diseases, but AFE now increases with forget set size. For instance, RMU forgetting 5\% of cancer patients achieves AFE below 0.2, but forgetting 25\% raises AFE to $\sim$0.6. So, \textbf{forget set size strongly influences RMU and SimNPO's unlearning effectiveness when patients share diseases.} This occurs because removing more patients from the same disease progressively erodes the shared medical knowledge, making it harder for the model to answer questions about that disease. 


In Figures \ref{fig:mu_cancer_overlay} and \ref{fig:mu_immune_overlay}, for RMU and SimNPO, MU on seen questions (averaged over factual and reasoning) for same-disease retained patients degrades 
substantially with forget set size: in the cancer cohort, both methods 
drop from ${\sim}0.83$ to ${\sim}0.37$, while in the infectious/immune 
cohort, RMU declines from ${\sim}0.87$ to ${\sim}0.62$ and SimNPO from ${\sim}0.84$ to ${\sim}0.53$. However, 
MU for out-of-disease patients remains high (${\sim}0.90$--$0.95$). 
GradDiff and KL-Min, by contrast, collapse to near-zero MU for 
\emph{same-disease} retained patients, but maintain moderate MU 
(${\sim}0.50$--$0.77$) for out-of-disease patients, indicating that their unlearning is disease-targeted rather than a global model collapse.

Holdout evaluation in Figure~\ref{fig:model_performance_disease} mirrors the trends observed for seen QAs: AFE increases with forget set size, and MU shows disease-dependent trajectories. \textbf{MU on reasoning QAs is consistently lower than MU on factual QAs, reflecting the greater difficulty of clinical inference.} Critically, holdout AFE values remain comparable to training AFE, unlike random patient unlearning, where holdout AFE values dropped to $\sim$0.2. This indicates that \textbf{disease-level patient unlearning operates at the patient-note level rather than just the QA level.}

\subsection{Do Models Reproduce Exact Medical Terms From Forgotten Patients?}

We now examine a novel privacy criterion: \textbf{disease keyword leakage}. Even if a model has high AFE, leakage of disease-related keywords from patient notes constitutes a privacy violation, as it confirms the model has retained specific information.

Table~\ref{tab:keyword_leakage} reports disease-specific keyword leakage rates on the forget set across disease categories and forget set sizes. GradDiff and KL-Min achieve near-zero leakage score (LS) ($\sim$0.001--0.007) across all diseases and forget set sizes, but this occurs because these methods generate largely incoherent outputs (\textbf{high degradation scores} in Table ~\ref{tab:keyword_leakage}) rather than because they have erased targeted information.

Disease-level forgetting shows a different pattern than random patient forgetting for RMU and SimNPO. For random patients, LS remains stable at $\sim$14--15\% (factual) and $\sim$25\% (reasoning) across all forget set sizes for RMU and SimNPO, consistent with the flat AFE patterns observed in Figure~\ref{fig:model_performance_patient}. In contrast, \textbf{for RMU and SimNPO, disease-level forgetting shows declining leakage as forget set size increases, with degradation remaining near zero across all diseases.} For example, for cancer, RMU's reasoning LS drops from 24.6\% (5\%) to 12.7\% (25\%), and SimNPO's drops from 24.1\% to 11.6\%. This mirrors the improving AFE trend in Figure~\ref{fig:model_performance_disease}, confirming that removing more patients from the same disease progressively erodes the shared medical knowledge, reducing leakage of medical terminology.

\section{Conclusion}

We introduced AMNESIA, a large-scale open source medical unlearning benchmark with 70,560 QAs from 8,820 patients, featuring random patient-level and disease-level splits. AMNESIA reveals the limitations of existing unlearning methods. For random patient-level splits, unlearning methods either fail to forget effectively or achieve forgetting through model collapse. Disease-level forgetting improves as the forget set grows, but at the cost of model utility on retained same-disease patients. Across both settings, clinical reasoning remains harder than factual recall. Holdout evaluation further shows that models forget specific QAs rather than the patient notes, since the underlying information remains accessible through alternative unseen questions. Our disease keyword leakage analysis also shows the importance of having a domain-specific question-answering evaluation metric in unlearning settings. Together, these findings highlight the need for medical-specific unlearning techniques that erase patient information without damaging shared medical knowledge and establish AMNESIA as an essential benchmark for advancing the field.

\section*{Limitations}

Our evaluation of forgetting relies on answer-level metrics (AFE, MU) and
exact-match keyword leakage. A model with high AFE may still encode patient
information recoverable through paraphrased queries, and our exact-match leakage metric does not catch paraphrased reproductions of forgotten
terminology. Therefore, what fraction of a patient's information is truly removed remains difficult to quantify.

All experiments use a single base model (\textsc{LLaMA 3-8B}), and gradient-based unlearning at layer ~7 to match the RMU configuration from OpenUnlearning. 

We partition patients into 11 single disease categories, but real clinical conditions exist on a spectrum, and comorbidities are common.

\section*{Ethical Considerations}

AMNESIA is built from publicly available, de-identified clinical notes and will be released without any restrictions on the source data.

Our central finding is that unlearning can coexist with substantial keyword
leakage, and new questions recover information from supposedly
forgotten patients. Deploying any of the methods we evaluate under the
assumption that they fulfill a right-to-be-forgotten request would create a
false sense of privacy compliance. We further show that forgetting a critical
mass of patients sharing a disease degrades model utility for retained
patients with the same condition, meaning batch deletion requests could
compromise care quality for patients who never requested deletion. None of the
models in this work is validated for clinical use.

\section*{Acknowledgments}

We thank our human annotator for volunteering to do our human-in-the-loop evaluation. 

We also acknowledge the use of Cursor \footnote{\url{https://cursor.com}} for coding assistance and Claude (Anthropic) \footnote{\url{https://www.anthropic.com/claude}} for writing assistance in preparing this manuscript.


\bibliography{custom}

\section*{Appendix}
\label{sec:appendix}
\appendix

\section{Comparison to Existing Benchmarks}
\label{app:prior_work}
\begin{table}[H]
\centering
\fontsize{9}{9}\selectfont
\setlength{\tabcolsep}{4pt}
\begin{tabular}{@{}lcccc@{}}
\toprule
\textbf{Benchmark} & \textbf{Modality} & \textbf{Scale} & \textbf{Disease} & \textbf{Grounded} \\
 & & & \textbf{Split} & \textbf{QA} \\
\midrule
\multicolumn{5}{l}{\textit{General Domain}} \\
\midrule
MUSE & Text & 20K & \xmark & \xmark \\
TOFU & Text & 4K & \xmark & \xmark \\
R-TOFU & Text & 4K & \xmark & \xmark \\
WMDP & Text & 3.6K & \xmark & \xmark \\
\midrule
\multicolumn{5}{l}{\textit{Medical Domain}} \\
\midrule
Med-Forget & Image+Text & 3.8K & \xmark & \xmark \\
Forget-MI & Image+Text & 6.7K & \xmark & \xmark \\
\midrule
\textbf{AMNESIA} & \textbf{Text} & \textbf{70.5K} & \cmark & \cmark \\
\textbf{(Ours)} \\
\bottomrule
\end{tabular}
\caption{Comparison of AMNESIA with existing unlearning benchmarks.}
\label{tab:benchmark-comparison}
\end{table}

\section{Prompts Used for Data Generation}
\label{app:prompts}

This appendix provides the complete prompts used in our automated data generation pipeline.

\subsection{Disease Category Assignment Prompt}
\label{app:disease-prompt}

We used the following prompt with \textsc{GPT-5.4-nano} to assign disease categories to patient cases based on article titles:

\begin{tcolorbox}[
    colback=gray!5,
    colframe=black!75,
    boxrule=0.5pt,
    arc=2pt,
    breakable,
    fontupper=\fontsize{9}{11}\selectfont
]
You are given a medical article with:
\begin{itemize}[noitemsep,topsep=2pt,leftmargin=*]
\item article\_id
\item article\_title
\end{itemize}

Your task is to assign exactly one top-level category to the article based primarily on the article title.

Use only this taxonomy:

\begin{enumerate}[noitemsep,topsep=2pt,leftmargin=*]
\item Anemia or Iron Deficiency
\item Arthritis and Bone
   \begin{itemize}[noitemsep,leftmargin=*]
   \item Arthritis
   \item Osteoporosis
   \end{itemize}
\item Cancer
\item Cardiovascular
   \begin{itemize}[noitemsep,leftmargin=*]
   \item Cerebrovascular Disease or Stroke
   \item Cholesterol
   \item Heart Disease
   \item Hypertension
   \end{itemize}
\item Dementia and Mental Health
   \begin{itemize}[noitemsep,leftmargin=*]
   \item Alzheimer's Disease
   \item Attention Deficit Hyperactivity Disorder
   \item Depression
   \item Mental Health
   \end{itemize}
\item Diabetes
\item Digestive and Liver
   \begin{itemize}[noitemsep,leftmargin=*]
   \item Digestive Diseases
   \item Chronic Liver Disease and Cirrhosis
   \end{itemize}
\item Infectious or Immune Diseases
   \begin{itemize}[noitemsep,leftmargin=*]
   \item AIDS and HIV
   \item Viral Hepatitis
   \item Infectious Disease
   \item Influenza
   \item Measles, Mumps, and Rubella
   \item Pneumonia
   \item Sexually Transmitted Diseases (STD)
   \item Chronic Sinusitis
   \item Whooping Cough or Pertussis
   \end{itemize}
\item Kidney Disease
\item Oral and Dental Health
\item Respiratory and Allergies
   \begin{itemize}[noitemsep,leftmargin=*]
   \item Allergies and Hay Fever
   \item Asthma
   \item Chronic Obstructive Pulmonary Disease
   \end{itemize}
\end{enumerate}

Rules:
\begin{itemize}[noitemsep,topsep=2pt,leftmargin=*]
\item Return the best matching top-level category only.
\item Base the decision mainly on the article\_title.
\item Use the subcategories only as guidance for mapping into the correct top-level category.
\item Do not use any category outside the taxonomy.
\item If no category is clearly supported by the title, choose ``Other'' and explain briefly.
\end{itemize}

Return valid JSON only in this exact format:
\begin{lstlisting}[language=json,basicstyle=\ttfamily\fontsize{9}{11}\selectfont,breaklines=true]
{
  "article_id": "<article_id>",
  "article_title": "<article_title>",
  "category": "<one top-level category from the taxonomy>",
  "short_reason": "<brief reason using title keywords>"
}
\end{lstlisting}
\end{tcolorbox}

\subsection{Question-Answer Generation Prompt}
\label{app:qa-prompt}

We used the following prompt with \textsc{GPT-5-nano} to generate question-answer pairs for each patient case:

\begin{tcolorbox}[
    colback=gray!5,
    colframe=black!75,
    boxrule=0.5pt,
    arc=2pt,
    breakable,
    fontupper=\fontsize{9}{11}\selectfont
]
You are a medical education expert creating a clinical knowledge assessment dataset.

Given a patient case summary, generate two groups of QA pairs:

\textbf{1. factual\_items}
\begin{itemize}[noitemsep,topsep=2pt,leftmargin=*]
\item direct recall from the summary
\item answerable solely from the provided summary
\item answers must be 1--2 sentences
\item cover diagnosis, symptoms, labs, treatments, outcomes
\end{itemize}

\textbf{2. reasoning\_items}
\begin{itemize}[noitemsep,topsep=2pt,leftmargin=*]
\item multi-hop clinical reasoning
\item each question MUST connect at least TWO findings from the summary
\item answers should explain the reasoning chain in 2--3 sentences
\end{itemize}

Global rules:
\begin{itemize}[noitemsep,topsep=2pt,leftmargin=*]
\item Do NOT ask about patient name, age, gender, or any PII
\item Questions must REQUIRE this specific patient's details and must NOT be answered confidently from general medical knowledge alone
\item Keep questions clinically specific and grounded
\item Return only valid JSON matching the schema
\end{itemize}
\end{tcolorbox}

\section{Data Statistics}
\label{app:data-statistics}
 
\begin{table}[!h]
\centering
\resizebox{\columnwidth}{!}{%
\fontsize{9}{9}\selectfont
\setlength{\tabcolsep}{2.5pt}
\begin{tabular}{@{}lrrrr@{}}
\toprule
\textbf{Disease Category} & \textbf{$|\mathrm{Q}|_{\mathrm{fact}}$} & \textbf{$|\mathrm{Q}|_{\mathrm{reas}}$} & \textbf{$|\mathrm{A}|_{\mathrm{fact}}$} & \textbf{$|\mathrm{A}|_{\mathrm{reas}}$} \\
\midrule
Cancer & 15.6 & 23.2 & 34.3 & 54.5 \\
Infectious/Immune & 15.6 & 22.8 & 34.5 & 55.7 \\
Cardiovascular & 16.1 & 23.3 & 34.6 & 56.1 \\
Digestive and Liver & 15.5 & 22.7 & 34.9 & 54.3 \\
Dementia and Mental Health & 15.4 & 23.8 & 31.4 & 55.0 \\
Arthritis and Bone & 15.1 & 22.4 & 36.1 & 53.3 \\
Respiratory and Allergies & 16.0 & 23.1 & 33.9 & 54.8 \\
Kidney Disease & 16.2 & 23.7 & 35.8 & 55.9 \\
Diabetes & 16.2 & 24.0 & 35.7 & 57.5 \\
Oral and Dental Health & 15.4 & 22.6 & 33.1 & 53.2 \\
Anemia or Iron Deficiency & 16.7 & 24.7 & 35.2 & 56.0 \\
\bottomrule
\end{tabular}}
\caption{Mean length (tokens) of questions ($|Q|$) and answers ($|A|$) in AMNESIA, by disease category.}
\label{tab:qa-length-by-category}
\end{table}

\begin{table}[!h]
\centering
\fontsize{9}{9}\selectfont
\begin{tabularx}{\columnwidth}{@{}rrrX@{}}
\toprule
\textbf{\%} & \textbf{\# Patients} & \textbf{\# QAs} & \textbf{Available For} \\
\midrule
\multicolumn{4}{@{}l}{\textit{Forget Splits}} \\
\midrule
5 & 441 & 2,646 & Random, Cancer, Inf/Imm., Cardio. \\
10 & 882 & 5,292 & Random, Cancer, Inf/Imm., Cardio. \\
15 & 1,323 & 7,938 & Random, Cancer, Inf/Imm., Cardio. \\
20 & 1,764 & 10,584 & Random, Cancer, Inf/Imm. \\
25 & 2,205 & 13,230 & Random, Cancer \\
\midrule
\multicolumn{4}{@{}l}{\textit{Retain Splits}} \\
\midrule
5 & 8,379 & 50,274 & Random, Cancer, Inf/Imm., Cardio. \\
10 & 7,938 & 47,628 & Random, Cancer, Inf/Imm., Cardio. \\
15 & 7,497 & 44,982 & Random, Cancer, Inf/Imm., Cardio. \\
20 & 7,056 & 42,336 & Random, Cancer, Inf/Imm. \\
25 & 6,615 & 39,690 & Random, Cancer \\
\midrule
\multicolumn{4}{@{}l}{\textit{Holdout Set}} \\
\midrule
-- & 8,820 & 17,640 & All \\
\bottomrule
\end{tabularx}
\caption{Data partition statistics. All forget/retain splits at each percentage contain identical patient/QA counts regardless of partition type.}
\label{tab:data-partition-full}
\end{table}

\section{Training Hyperparameters}
\label{app:training-hyperparams}
 
We perform pre-training on patient clinical notes to adapt \textsc{LLaMA 3-8B} model to medical language and terminology. Patient notes are tokenized using the \textsc{LLaMA 3} tokenizer with a maximum sequence length of 2048 tokens and then packed to maximize GPU utilization. We employ distributed training across 8 AMD MI250X GPUs. Gradient checkpointing is enabled to reduce memory consumption. Unlearning also occurred for 3 epochs, and all experiments are from the last checkpoint of each unlearned model (the same as finetuning).

\begin{table}[h!]
\centering
\fontsize{9}{11}\selectfont
\begin{tabularx}{\columnwidth}{@{}lXX@{}}
\toprule
\textbf{Hyperparameter} & \textbf{Pretraining} & \textbf{Finetuning} \\
\midrule
\multicolumn{3}{@{}l}{\textit{Model \& Data}} \\
Base model & \textsc{LLaMA 3-8B} & Pretrain checkpoint \\
Training data & 8820 notes & notes + $\sim$52.9K QAs \\
Max sequence length & 2048 & 2048 \\
\midrule
\multicolumn{3}{@{}l}{\textit{Optimization}} \\
Optimizer & AdamW & AdamW \\
Learning rate & $2e\text{-}5$ & $2e\text{-}5$ \\
LR scheduler & Cosine & Cosine \\
Warmup ratio & 0.03 & 0.03 \\
Weight decay & 0.0 & 0.0 \\
Epochs & 1 & 3 \\
\midrule
\multicolumn{3}{@{}l}{\textit{Batching}} \\
Per-device batch size & 2 & 2 \\
Gradient accumulation steps & 4 & 4 \\
Effective batch size & 64 & 64 \\
\midrule
\multicolumn{3}{@{}l}{\textit{Infrastructure}} \\
GPUs & 8 MI250X & 8 MI250X \\
Gradient checkpoint & On & On \\
\bottomrule
\end{tabularx}
\caption{Training hyperparameters for pretraining and instruction finetuning.}
\label{tab:training-hyperparams}
\end{table}

During instruction fine-tuning, we use the following template to structure each training example:
 
\begin{tcolorbox}[
    colback=gray!5,
    colframe=black!75,
    boxrule=0.5pt,
    arc=2pt,
    fontupper=\fontsize{9}{11}\selectfont,
    breakable
]
\texttt{You are an intelligent clinical language model.}
 
\texttt{Below is a snippet of patient's discharge summary and the following instruction from a healthcare professional. Write a response that appropriately completes the instruction. The response should provide the accurate answer to the instruction, while being concise.}
 
\texttt{[Discharge Summary Begin]}
 
\texttt{\{patient\_note\}}
 
\texttt{[Discharge Summary End]}
 
\texttt{[Instruction Begin]}
 
\texttt{\{question\}}
 
\texttt{[Instruction End]}
\end{tcolorbox}

\section{Leakage Score (LS)}
\label{app:ls}

We introduce a disease-specific keyword extraction pipeline to detect whether unlearned models continue reproducing specific medical terminology from forget patients. This provides fine-grained evaluation beyond aggregate metrics.
 
\subsection{Extraction Pipeline}
 
For each patient in the dataset, we extract disease-relevant keywords from three text sources: patient notes, questions, and answers. The pipeline uses scispaCy \citep{neumann2019scispacy} medical NLP models to identify biomedical entities and noun chunks. Specifically, we use en\_core\_sci\_lg for patient notes, a large English biomedical model trained on scientific literature with word2vec embeddings that can process arbitrarily long texts without token-length restrictions. For questions and answers, the same model is used by default, though the pipeline supports alternative models such as en\_core\_sci\_scibert (a BERT-based transformer model with higher run-time and a 512-token limit). We focus on three major disease categories: Cancer, Cardiovascular, and Infectious or Immune Diseases.

\begin{table*}[t]
\centering
\footnotesize
\begin{tabularx}{\textwidth}{@{}l l X@{}}
\toprule
\textbf{Category} & \textbf{Subcategory} & \textbf{Terms} \\
\midrule
Cancer & General oncology    & cancer, tumor, tumour, neoplasm, malignancy, malignant, carcinoma, sarcoma, lymphoma, leukemia, leukaemia, myeloma, glioma, blastoma \\
       & Pathology           & metastasis, metastatic, adenocarcinoma, melanoma, squamous cell, hepatocellular, biopsy, cytology, grade, staging, margins, excision \\
       & Specific cancers    & lung cancer, breast cancer, colon cancer, prostate cancer, ovarian cancer, pancreatic cancer, lymph node, bone marrow, recurrence, relapse \\
       & Treatment           & chemotherapy, radiotherapy, radiation therapy, adjuvant, neoadjuvant, palliative, targeted therapy, hormone therapy, resection, debulking, pet scan, screening, remission, immunotherapy \\
\midrule
Cardiovascular & General    & cardiovascular, cardiac, coronary, myocardial, infarction, angina, arrhythmia, atrial fibrillation, ventricular, heart failure, cardiomyopathy, pericarditis \\
               & Conditions & hypertension, atherosclerosis, ischemia, ischaemia, stroke, aortic, mitral, valve, embolism, thrombosis, dvt, pulmonary embolism, bradycardia, tachycardia, syncope, heart attack \\
               & Biomarkers & troponin, bnp, ntprobnp, ejection fraction, lipids, cholesterol, statin \\
               & Procedures & stent, cabg, bypass, angioplasty, revascularization, echocardiogram, ecg, ekg, defibrillation, pacemaker, catheterization, chest pain, cad, chf, hfref, hfpef \\
\midrule
Infectious or Immune & Infection (general) & infection, infectious, sepsis, bacteremia, fever, abscess, cellulitis, wound infection \\
                     & Pathogens           & viral, virus, bacterial, bacteria, fungal, candida, pathogen, mrsa, tuberculosis, covid, hiv, hepatitis, influenza \\
                     & Specific infections & pneumonia, meningitis, encephalitis, osteomyelitis, uti \\
                     & Immune system       & immune, immunity, autoimmune, immunosuppression, immunocompromised, complement, cytokine, interferon, rheumatoid, lupus, vasculitis, seropositive \\
                     & Inflammation markers & inflammation, inflammatory, crp, esr, leukocytosis, neutrophil, lymphocyte, wbc \\
                     & Diagnostics \& treatment & antibiotic, antiviral, antifungal, prophylaxis, culture, pcr, serology \\
\midrule
Common Terms & --- & patient, patients, year, years, old, male, female, history, medical history, past medical history, pmh, today, yesterday, day, days, week, weeks, month, months, time, normal, none, no, yes, denies, reported, reports, present, mg, ml, cm, right, left, bilateral, mild, moderate, severe, acute, chronic, follow, follow up, plan, assessment, exam, physical exam, question, answer, following, option, options, true, false \\
\bottomrule
\end{tabularx}
\caption{Disease-specific seed terms used for filtering, organized by
category and subcategory, plus the common-clinical term list.}
\label{tab:seed-terms}
\end{table*}

\paragraph{Disease-Specific Seed Terms}
For each disease category, we maintain a curated list of seed terms (Table \ref{tab:seed-terms}) covering core terminology, conditions, biomarkers, and treatments. These seed lists were curated by prompting Claude (Anthropic) to generate medical terminology for each disease category, followed by manual review and validation by the authors. We also filter out common clinical terms that lack patient-specific diagnostic value.

\begin{algorithm}[h]
\caption{Disease Keyword Extraction and Leakage Score Calculation}
\label{alg:keyword-leakage}
\fontsize{9}{11}\selectfont
\begin{algorithmic}[1]
\Require Patient data $(note, question, answer)$, disease category $d$, seed terms $S_d$
\Ensure Keyword leakage score for prediction $y$
 
\State \textbf{// Step 1: Extract candidate phrases}
\State $candidates \gets \emptyset$
\For{each text in $\{note, question, answer\}$}
    \State $doc \gets$ scispaCy.parse($text$)
    \State $candidates \gets candidates \cup$ ExtractEntities($doc$)
    \State $candidates \gets candidates \cup$ ExtractNounChunks($doc$)
\EndFor
\State Filter out generic terms, short ($<$3 char) and long ($>$6 words) phrases
 
\State \textbf{// Step 2: Score phrase relevance to disease}
\State $scored \gets []$
\For{each phrase $p$ in $candidates$}
    \State $rel \gets$ ComputeRelevance($p$, $S_d$) \Comment{Score 0.0--1.0}
    \If{$rel \geq 0.5$}
        \State $freq \gets$ Count($p$ in $candidates$)
        \State $score \gets rel + \min(freq, 5) \times 0.05$
        \State Append $(p, score)$ to $scored$
    \EndIf
\EndFor
 
\State \textbf{// Step 3: Select top-K and remove redundancy}
\State Sort $scored$ by score (descending)
\State $keywords \gets$ SelectTopK($scored$, $k=5$ for notes, all for Q\&A)
\State $keywords \gets$ RemoveRedundant($keywords$) \Comment{Keep shorter phrases}
 
\State \textbf{// Step 4: Compute leakage on forget-set predictions}
\State $y \gets$ Model.predict($question$) \Comment{Prediction for patient $p$}
\State Tokenize and lowercase $y$: $y_{tokens} \gets$ Tokenize($y$)
\State $numerator \gets 0$, $denominator \gets 0$
\For{each $(k_i, s_i)$ in $keywords$}
    \State $k_{tokens} \gets$ Tokenize($k_i$)
    \State $match \gets$ IsContiguousSubsequence($k_{tokens}$, $y_{tokens}$)
    \State $numerator \gets numerator + s_i \times match$
    \State $denominator \gets denominator + s_i$
\EndFor
\State \Return $\frac{numerator}{denominator}$ \Comment{Leakage score $\in [0, 1]$}
\end{algorithmic}
\end{algorithm}

\paragraph{Candidate Extraction}
From each scispaCy-parsed document, we extract:
\begin{enumerate}[noitemsep,topsep=2pt,leftmargin=*]
\item Named entities identified by the biomedical entity recognizer
\item Noun chunks from the dependency parser
\item Filtering: remove common clinical terms, phrases $<$ 3 characters, and phrases $>$ 6 words
\end{enumerate}
 
\paragraph{Relevance Scoring}
Each candidate phrase receives a lexical disease relevance score based on its overlap with seed terms:
\begin{itemize}[noitemsep,topsep=2pt,leftmargin=*]
\item \textbf{Exact match:} score = 1.0
\item \textbf{Multi-word phrase containment:} score = 0.9 (e.g., ``lung cancer'' contains seed ``lung cancer'')
\item \textbf{Partial word overlap:} score = 0.0--0.6 (proportional to Jaccard similarity)
\end{itemize}
Phrases scoring below 0.5 are filtered. The final score combines relevance with frequency: $\text{score} = \text{relevance} + \min(\text{freq}, 5) \times 0.05$.
 
\paragraph{Top-K Selection}
For patient notes, we select the top-5 highest-scoring keywords after removing redundant longer phrases that contain shorter selected keywords (e.g., keep ``lung cancer'', remove ``small tumor lung cancer''). For questions and answers, we extract all keywords meeting the relevance threshold without top-k limitation, as these texts are shorter and contain fewer candidates.
 
\subsection{Keyword Leakage Computation}
 
For each model prediction $y$ on a forget-set question, we compute the weighted keyword leakage score:
\begin{equation}
\text{Leakage}(y) = \frac{\sum_{i} s_i \cdot \mathbb{1}[k_i \in y]}{\sum_{i} s_i}
\end{equation}
where $k_i$ are the extracted keywords from the corresponding patient (note + question + answer), $s_i$ are their relevance scores, and $\mathbb{1}[k_i \in y]$ indicates the exact phrase match detected via tokenized n-gram comparison. The prediction and keywords are lowercased and tokenized (alphanumeric tokens only), then we check if each keyword appears as a contiguous sequence in the prediction tokens.
 
This metric quantifies the extent to which the model reproduces disease-specific medical terminology associated with forget patients. Algorithm \ref{alg:keyword-leakage} shows the entire pipeline.

\section{Human-in-the-Loop Validation of Disease Category Labels}
\label{app:human-validation}

This appendix details the LLM-as-judge panel used to verify disease
category assignments in AMNESIA, together with the human validation study used to verify the panel itself.

\subsection{Panel Design and Models}
\label{app:panel-design}

We assembled a three-judge panel with deliberately heterogeneous families to limit single-model bias:

\begin{itemize}[leftmargin=*,itemsep=2pt,topsep=2pt]

  \item \textbf{MedGemma-27B-IT} (\texttt{google/medgemma-27b-it}):
        A 27B-parameter medical-specialized instruction-tuned model by Google
        DeepMind, built on the Gemma~3 backbone and trained on clinical text,
        medical QA pairs, and EHR data. It achieves 87.7\% on MedQA, making it
        one of the strongest open-weight models for clinical reasoning and
        medical question answering.

  \item \textbf{Qwen3-32B} (\texttt{Qwen/Qwen3-32B}):
        A 32.8B-parameter open-weight model by Alibaba released under Apache~2.0,
        featuring a hybrid architecture that switches between extended
        chain-of-thought \emph{thinking mode} and fast \emph{non-thinking mode}.
        It supports a 128K-token context window with strong performance in
        mathematics, coding, and agentic reasoning across 119 languages.

  \item \textbf{GPT-5 mini} (\texttt{gpt-5-mini-2025-08-07}):
        A closed-source proprietary model by OpenAI and the cost-efficient
        variant of the flagship GPT-5 (released August 2025), supporting a
        272K-token context window with vision, function calling, and web search.
        It delivers approximately 45\% fewer hallucinations than GPT-4o while
        targeting high-volume, low-latency workloads requiring reliable
        multi-step reasoning.

\end{itemize}

Each judge received the article title and assigned category and was
required to return a JSON object with three fields:
(i)~a 1--2 sentence rationale, (ii)~a verdict in
$\{\textit{Supported}, \textit{Unsupported}, \textit{Ambiguous}\}$,
and (iii)~the category the judge believed best fit the title, drawn
from \{Cardiovascular, Cancer, Infectious/Immune Diseases, Unclear\}. The final per-title label was the
majority verdict across the three judges; ties (one judge each on three
different labels) were recorded as \textit{Disagree}. 

\begin{tcolorbox}[
    colback=gray!5,
    colframe=black!75,
    boxrule=0.5pt,
    arc=2pt,
    breakable,
    fontupper=\fontsize{9}{11}\selectfont,
    label=lst:judge-prompt
]
\textbf{[System]}\\
You are a medical research expert evaluating whether a category label
is correctly assigned to a medical article title.
\medskip

The three valid categories and their definitions:\\
- Cardiovascular: heart disease, vascular conditions, blood pressure,
cardiac interventions, stroke (cerebrovascular), arrhythmias, coronary
artery disease.\\
- Cancer: oncology, tumors, malignancies, carcinomas, chemotherapy,
radiotherapy, cancer screening, neoplasms.\\
- Infectious or Immune Diseases: infections (bacterial, viral, fungal,
parasitic), vaccines, autoimmune conditions, immunodeficiency,
inflammatory immune responses, sepsis.
\medskip

You must respond ONLY with a valid JSON object.
\medskip

\textbf{[User]}\\
Article Title: ``\{title\}''\\
Assigned Category: ``\{category\}''
\medskip

Task: Does the assigned category accurately reflect the medical focus
of this title?
\medskip

Reason step by step:\\
1. What is the primary medical topic of this title?\\
2. Which of the three categories does it best fit?\\
3. Does the assigned category match?
\medskip

Respond ONLY with this JSON (no markdown, no extra text):\\
\{\\
\ \ "reasoning": "<1-2 sentence explanation>",\\
\ \ "verdict": "<Supported | Unsupported | Ambiguous>",\\
\ \ "correct\_category": "<Cardiovascular | Cancer | Infectious or
Immune Diseases | Unclear>"\\
\}
\end{tcolorbox}

\subsection{Panel Results on 900 Titles}
\label{app:panel-results}

The panel evaluated all 900 sampled patient titles spanning Cancer
(300 titles), Cardiovascular (300 titles), and Infectious/Immune
Diseases (300 titles). 
Table~\ref{tab:verdict-distribution} shows the verdict distribution across models. Table~\ref{tab:panel-by-cat}
breaks the panel verdict down by assigned disease, and
Table~\ref{tab:pairwise-judges} reports pairwise inter-judge agreement.

\begin{table}[h]
\centering
\setlength{\tabcolsep}{4pt}
\resizebox{\columnwidth}{!}{%
\begin{tabular}{lrrrr}
\toprule
\textbf{Verdict} & \textbf{MedGemma} & \textbf{Qwen} & \textbf{GPT-5-m} & \textbf{Panel} \\
\midrule
Supported   & 819 (91.0\%) & 827 (91.9\%) & 827 (91.9\%) & 826 (91.8\%) \\
Unsupported &  80 (8.9\%)  &  65 (7.2\%)  &  70 (7.8\%)  &  71 (7.9\%)  \\
Ambiguous   &   1 (0.1\%)  &   8 (0.9\%)  &   3 (0.3\%)  &   1 (0.1\%)  \\
Disagree    &   ---        &   ---        &   ---        &   2 (0.2\%)  \\
\bottomrule
\end{tabular}}
\caption{Verdict distribution across individual judges and the panel consensus. Judges show strong agreement on \textit{Supported} cases ($\sim$92\%), with minor differences in how borderline cases are split between \textit{Unsupported} and \textit{Ambiguous}.}
\label{tab:verdict-distribution}
\end{table}

\begin{table}[h]
\centering
\setlength{\tabcolsep}{2pt}
\begin{tabular}{lrrrr}
\toprule
\textbf{Disease} & \textbf{Supp.} & \textbf{Unsupp.} & \textbf{Amb.} & \textbf{Dis.} \\
\midrule
Cancer                          & 284 & 16 & 0 & 0 \\
Cardiovascular                  & 283 & 16 & 1 & 0 \\
Infectious / Immune             & 259 & 39 & 0 & 2 \\
\midrule
\textbf{Total}                  & 826 & 71 & 1 & 2 \\
\bottomrule
\end{tabular}
\caption{Panel verdict by assigned disease. The Infectious/Immune
disease carries the largest fraction of \textit{Unsupported} cases,
reflecting the breadth of its definition (bacterial, viral, fungal,
parasitic, autoimmune, immunodeficiency) and the higher chance of
titles co-occurring with cardiovascular or oncologic primary topics.}
\label{tab:panel-by-cat}
\end{table}

\begin{table}[h]
\centering
\begin{tabularx}{\columnwidth}{@{}Xr@{}}
\toprule
\textbf{Pair} & \textbf{Agreement} \\
\midrule
MedGemma $\leftrightarrow$ Qwen        & 95.6\% \\
MedGemma $\leftrightarrow$ GPT-5-mini  & 95.1\% \\
Qwen $\leftrightarrow$ GPT-5-mini      & 95.8\% \\
\midrule
\textbf{Fleiss' $\kappa$ (binary)}     & \textbf{0.726} (Substantial) \\
\textbf{Avg.\ per-row agreement}       & \textbf{95.5\%} \\
\bottomrule
\end{tabularx}
\caption{Pairwise inter-judge agreement and Fleiss' $\kappa$
(collapsing \textit{Unsupported}/\textit{Ambiguous} against
\textit{Supported}) over $N=900$ titles.}
\label{tab:pairwise-judges}
\end{table}

\subsection{Human Validation Protocol}
\label{app:human-protocol}

To evaluate the panel, one biology student manually verified the aforementioned 900
titles. For each title, the annotator was shown the title and the assigned category and answered a
binary question: \emph{``Is the assigned category label supported by
the title?''}\ (Yes/No). The annotator was blinded to all LLM judge
outputs and to the panel verdict at annotation time. The resulting
human label distribution was $870$ Yes and $30$ No.

We aligned the human Yes/No labels with the LLM \textit{Supported}/
\textit{Unsupported} scheme and computed agreement statistics
per judge against the human. We report a confusion
matrix per judge (Table~\ref{tab:human-vs-judges}). For each judge's confusion matrix, that judge's Ambiguous verdicts were excluded from the denominator (1 for MedGemma, 8 for Qwen, 3 for GPT-5-mini). For the panel matrix, the 1 Ambiguous and 2 Disagree panel outcomes were excluded, yielding 897 rows.




\section{LLM-as-Judge Validation of QA Pairs}
\label{app:qa-validity}

We sampled two QAs (one factual and one reasoning) per 500 randomly selected patients, resulting in $1{,}000$ \emph{(patient note, question, answer)} triples from the AMNESIA training pool.  We submitted each to the same
three-judge panel used for disease-label validation
(Appendix~\ref{app:human-validation}): \textsc{MedGemma-27B},
\textsc{Qwen3-32B},and \textsc{GPT-5-mini}.

\begin{table}[h]
\centering
\setlength{\tabcolsep}{14pt}
\begin{tabular}{l|cc}
\multicolumn{3}{l}{\textbf{(a) MedGemma}} \\
\toprule
human $\backslash$ judge & Supp. & Unsupp. \\
\midrule
Supported   & 805 & 64 \\
Unsupported &   14 & 16 \\
\bottomrule
\end{tabular}
\quad
\begin{tabular}{l|cc}
\multicolumn{3}{l}{\textbf{(b) Qwen}} \\
\toprule
human $\backslash$ judge & Supp. & Unsupp. \\
\midrule
Supported   & 813 & 52 \\
Unsupported &   14 & 13 \\
\bottomrule
\end{tabular}
\vspace{4pt}
\begin{tabular}{l|cc}
\multicolumn{3}{l}{\textbf{(c) GPT-5-mini}} \\
\toprule
human $\backslash$ judge & Supp. & Unsupp. \\
\midrule
Supported   & 816 & 52 \\
Unsupported &   11 & 18 \\
\bottomrule
\end{tabular}
\quad
\begin{tabular}{l|cc}
\multicolumn{3}{l}{\textbf{(d) Panel}} \\
\toprule
human $\backslash$ judge & Supp. & Unsupp. \\
\midrule
Supported   & 814 & 54 \\
Unsupported &   12 & 17 \\
\bottomrule
\end{tabular}
\caption{Confusion matrices for human (rows) vs.\ each LLM judge and the panel (columns) on the 900 human-annotated titles.}
\label{tab:human-vs-judges}
\end{table}

\subsection{Judge Prompt}
\label{app:judge-prompt}

Each judge was given the full patient note alongside the candidate
question and answer, and asked for a strict binary verdict together
with a brief rationale. The verbatim prompt is below.

\begin{tcolorbox}[
    colback=gray!5,
    colframe=black!75,
    boxrule=0.5pt,
    arc=2pt,
    breakable,
    fontupper=\fontsize{9}{11}\selectfont,
    label=lst:qa-judge-prompt
]
\textbf{[System]}\\
You are a medical AI expert evaluating the quality of question-answer
pairs derived from clinical case reports.\\
You must respond ONLY with a valid JSON object.
\medskip

\textbf{[User]}\\
You are evaluating the quality of a medical question-answer pair.
\medskip

PATIENT NOTE:\\
\{patient\_note\}
\medskip

QUESTION:\\
\{question\}
\medskip

ANSWER:\\
\{answer\}
\medskip

TASK:\\
Evaluate if this QA pair is valid for a medical AI benchmark. A valid
QA pair must meet ALL criteria:\\
1. The question is clear and well-formed\\
2. The answer is correct based on the patient note\\
3. The answer can be derived from the information in the patient note\\
4. The question and answer are clinically meaningful
\medskip

Provide your evaluation in JSON format:\\
\{\\
\ \ "valid": true/false,\\
\ \ "reasoning": "Brief explanation of your decision (2-3 sentences)"\\
\}
\medskip

Be strict in your evaluation. If ANY criterion is not met, mark as
invalid.
\end{tcolorbox}

\subsection{Panel Results}
\label{app:qa-results}

Table~\ref{tab:qa-panel} summarizes verdicts at the judge and panel
level over the $N=1{,}000$ triples, and Table~\ref{tab:qa-pairwise}
reports pairwise inter-judge agreement and Fleiss' $\kappa$. 
\begin{table}[!h]
\centering
\resizebox{\columnwidth}{!}{%
\begin{tabular}{@{}lrrrr@{}}
\toprule
\textbf{Verdict} & \textbf{MedGemma} & \textbf{Qwen} & \textbf{GPT-5-m} & \textbf{Panel} \\
\midrule
Valid    & 957 (95.7\%) & 970 (97.0\%) & 952 (95.2\%) & 976 (97.6\%) \\
Invalid  &  42 (4.2\%)  &  29 (2.9\%)  &  48 (4.8\%)  &  23 (2.3\%)  \\
Unknown  &   1 (0.1\%)  &   1 (0.1\%)  &   0 (0.0\%)  &   ---        \\
Split    &   ---        &   ---        &   ---        &   1 (0.1\%)  \\
\bottomrule
\end{tabular}}
\caption{QA-validity verdict distribution across $N=1{,}000$
triples. The panel verdict is the majority of the three judges; a
single three-way split case (one judge \textit{Unknown}, one
\textit{Valid}, one \textit{Invalid}) is reported as
\textit{Split}.}
\label{tab:qa-panel}
\end{table}

\begin{table}[H]
\centering
\begin{tabularx}{\columnwidth}{@{}Xr@{}}
\toprule
\textbf{Pair} & \textbf{Agreement} \\
\midrule
MedGemma $\leftrightarrow$ Qwen        & 94.3\% \\
MedGemma $\leftrightarrow$ GPT-5-mini  & 93.6\% \\
Qwen $\leftrightarrow$ GPT-5-mini      & 93.8\% \\
\midrule
\textbf{Avg.\ per-row agreement}        & \textbf{93.9\%} \\
\textbf{Fleiss' $\kappa$} (\textit{Valid} vs.\ \textit{Invalid}) & \textbf{0.216} (Fair) \\
\bottomrule
\end{tabularx}
\caption{Pairwise inter-judge agreement and Fleiss' $\kappa$ for
QA-pair validation. The low $\kappa$ reflects skewed prevalence
($\approx 96\%$ \textit{Valid}) rather than substantive
disagreement; raw agreement remains high.}
\label{tab:qa-pairwise}
\end{table}

\end{document}